\documentclass[runningheads]{llncs}

\usepackage{booktabs}
\usepackage{multirow}
\usepackage{amsmath}
\usepackage{amssymb}
\usepackage{graphicx}
\usepackage{algorithm}
\usepackage[noend]{algpseudocode}
\usepackage[colorlinks=true,linkcolor=blue,citecolor=blue]{hyperref}

\begin{document}

\title{
    OrthoSkillVLA: Continual Skill Learning via Gradient-Informed Skill Subspace Adaptation
}
\titlerunning{OrthoSkillVLA}

\author{
    Jiaqi Wang\inst{1,2} \and
    Zhou Fang\inst{1,2} \and
    Qiongfeng Shi\inst{3} \and
    Yi Zhou\inst{1,2}\thanks{Corresponding author.}
}
\authorrunning{J. Wang and Y. Zhou et al.}

\institute{
    School of Computer Science and Engineering, Southeast University, China \and
    Key Laboratory of New Generation Artificial Intelligence Technology and Its Interdisciplinary Applications, Ministry of Education, China \and
    School of Electronic Science and Engineering, Southeast University, China
    \\
    \email{\{jiaqi.wang, 220245032, qiongfeng\}@seu.edu.cn, yizhou.szcn@gmail.com}
    \url{https://github.com/Jiaqi-Wangx/OrthoSkillVLA}
}

\maketitle

\begin{abstract}
Pretrained Vision-Language-Action models provide a strong foundation for robot learning, but sequentially adapting them to diverse skills can perturb the representations and velocity mappings used by previous skills, leading to catastrophic forgetting.
Architecture-based approaches improve retention by isolating skills but lead to increased inference footprint.
Recent subspace-constrained methods restrict parameter updates in an orthogonal subspace to minimize interference but impose a unified constraint on the entire model.
We analyze the distinct roles of internal VLA components and identify two VLA-specific challenges.
First, the VLM maintains broad semantic representations, making it vulnerable to capacity exhaustion, whereas the ActionHead refines semantics into localized velocity patterns that are highly sensitive to perturbations.
Second, the final velocity decoder serves as a readout layer.
Freezing it forms an output-stage expressivity bottleneck, while updating it risks overwriting previous velocity mappings.
To this end, we propose \textbf{OrthoSkillVLA}, a parameter-efficient framework for continual skill learning in pretrained VLA models without demonstration replay.
Given the representation heterogeneity, we impose separate subspace constraints on the VLM and ActionHead, preserving reusable semantic capacity while protecting localized velocity patterns.
For the output layer, we introduce a lightweight feature-aware MoE decoder, where each skill is allocated a compact expert and a training-free router selects the expert according to feature-space affinity.
Extensive simulated and real-world evaluations, together with ablations, demonstrate that OrthoSkillVLA better preserves prior skills while acquiring new ones.

\end{abstract}

\section{Introduction}
Recent Vision-Language-Action (VLA) models are pretrained to predict conditional velocity fields, enabling them to capture diverse and multimodal action distributions~\cite{intelligence2025pi05,bjorck2025gr00t}.
While these models exhibit strong multi-task adaptation capabilities~\cite{zheng2025xvla}, they still struggle to continually acquire skills~\cite{zheng2025imanip}.
In fitting significantly varying behavioral patterns, the parameter updates can interfere with established representations of previously learned skills, leading to distorted velocity-field predictions and severe performance degradation.
This challenge compromises the scalable deployment of VLA models, motivating continual learning frameworks that support lifelong skill acquisition~\cite{zhang2026atomicvla,liu2026pretrainedResistant}.

Existing architecture-based approaches alleviate forgetting by isolating skills into separate computational paths~\cite{romer2026clare,wu2025stellarvla,zhang2026atomicvla}.
However, such isolation often couples retention with skill-dependent architectural growth.
As skills accumulate, these added components introduce a growing footprint and auxiliary computation, which can be undesirable for resource-constrained real-time robotic control~\cite{fang2026probeflow}.
Gradient-projection-based methods constrain parameter updates to subspaces approximately orthogonal to previously occupied directions, thereby protecting existing knowledge~\cite{luo2026keeplora,gu2026eblora}.
These subspace-constrained methods maintain a constant architectural footprint after continual skill acquisition, offering a highly practical alternative for real-world deployment of VLA models.

However, directly transplanting subspace-constrained methods to VLA models is insufficient for two VLA-specific reasons.
First, existing methods typically impose uniform constraints across the network, failing to account for the distinct roles of three core components of VLA architectures: the Vision-Language Model (VLM), the ActionHead and the final velocity decoder.
The VLM encodes reusable semantic representations across skills, whereas the ActionHead operates on more localized velocity patterns, where even small perturbations can corrupt the final action predictions.
Empirically, this representational heterogeneity induces an inherent trade-off: a stringent constraint required by the ActionHead rapidly consumes the VLM's finite adaptation capacity, while a relaxed one that preserves the VLM's plasticity fails to protect delicate velocity patterns.
Such a mismatch is largely absent in vision or language continual learning~\cite{wang2022dualprompt,wang2023olora,luo2025lada}, but becomes critical for VLA models where semantic grounding and velocity generation are tightly coupled.
Second, the final velocity decoder is not simply another hidden layer but a critical output stage.
As the final readout from latent representations to velocity fields, updating the shared decoder can directly overwrite previously learned velocity mappings.
Freezing it avoids such disruption, but shifts the burden of fitting heterogeneous velocity fields onto the preceding adapters.
Since these adapters are themselves confined to orthogonal subspaces for stability, forcing them to compensate for a rigid output layer further intensifies the stability-plasticity tension.

In this work, we propose \textbf{OrthoSkillVLA}, a parameter-efficient continual learning framework for pretrained VLA models.
OrthoSkillVLA estimates skill feature subspaces from accumulated gradients and injects new skill knowledge into the orthogonal complement of previously occupied directions.
Crucially, rather than uniformly constraining all modules, this orthogonal adaptation is applied in a module-aware manner.
We assign separate subspace budgets to the VLM and the ActionHead, thereby balancing long-term semantic plasticity with high-precision velocity pattern retention.
For the final velocity decoder, which forms an output-stage expressivity bottleneck, we exclude it from orthogonal fine-tuning and introduce a lightweight Feature-Aware MoE Decoder.
Each new skill is assigned a compact expert to preserve velocity-field expressivity, while a training-free projection-based router selects experts according to feature-space affinity derived from gradient-informed skill bases.
Our contributions are summarized as follows:
\begin{itemize}
    \item We conduct empirical studies and reveal two specific challenges of pretrained VLA models: module-level representational heterogeneity and output-stage expressivity bottleneck.
    \item We introduce a module-level subspace budgeting strategy, enabling effective skill knowledge retention while preserving semantic plasticity for subsequent skill learning. For the velocity decoder, we design a Feature-Aware MoE Decoder with a training-free projection-based routing mechanism, which maintains velocity decoding expressivity with minimal architectural overhead.
    \item We validate OrthoSkillVLA on simulated and real-world environments, showing consistent improvements over subspace-constrained baselines in prior-skill retention and final success rates.
\end{itemize}

\section{Related Work}
\subsection{Pretrained Vision-Language-Action Models}
Vision-Language-Action (VLA) models pretrained on large-scale datasets~\cite{o2024openxembodiment,lim2024droid} serve as powerful foundations for end-to-end robotic control~\cite{kim2025openvla,zitkovich2023rt,bu2025agibot_iros}.
Modern VLA models have increasingly converged on flow-matching paradigms~\cite{lipmanflowmatching}.

By predicting continuous velocity fields, pioneering models like $\pi_{0.5}$~\cite{intelligence2025pi05} and GR00T~\cite{bjorck2025gr00t} successfully capture highly complex action distributions, resulting in precise and highly coherent trajectory generation.
Driven by these robust generative objectives, recent advancements such as X-VLA~\cite{zheng2025xvla} and SmolVLA~\cite{shukor2025smolvla} have successfully introduced streamlined architectures and structural miniaturizations.
These compact yet capable models significantly alleviate the computational burden on physical platforms, providing ideal testbeds for real-world deployment.
However, maintaining this rigid architectural efficiency while sequentially acquiring diverse skills remains a formidable challenge.

\subsection{Continual Learning for Robotics and VLA Models}
Continual learning in robotic manipulation aims to enable embodied agents to sequentially acquire new skills without forgetting previously mastered ones.
Prior robotic continual learning methods mainly rely on replay, skill libraries, or hierarchical skill representations~\cite{zheng2025imanip,wan2024lotus,xu2025speci}, whereas our work focuses on replay-free continual adaptation of pretrained flow-matching VLA models.
Recent VLA-oriented approaches further mitigate forgetting through adapter expansion~\cite{romer2026clare}, expert routing, evolving skill spaces~\cite{wu2025stellarvla}, or atomic skill decomposition~\cite{zhang2026atomicvla}.
While these mechanisms improve skill isolation, they often require explicit routing modules, growing skill-specific components, or additional skill decomposition procedures, which may increase structural complexity and inference overhead as skills accumulate.

Another line of work mitigates forgetting by constraining the optimization trajectory rather than expanding model architecture.
Gradient-projection and subspace-constrained methods estimate important feature or gradient subspaces of prior tasks~\cite{saha2021gpm} and restrict subsequent updates to low-interference directions~\cite{wang2023olora,liang2024inflora,luo2026keeplora,gu2026eblora}.
These methods provide a natural foundation for parameter-efficient continual adaptation, but their model-level or unified constraints do not directly account for the module-level heterogeneity and the low-dimensional velocity decoding bottleneck of flow-matching VLA policies.

\begin{figure}[t]
    \centering
    \includegraphics[width=\textwidth]{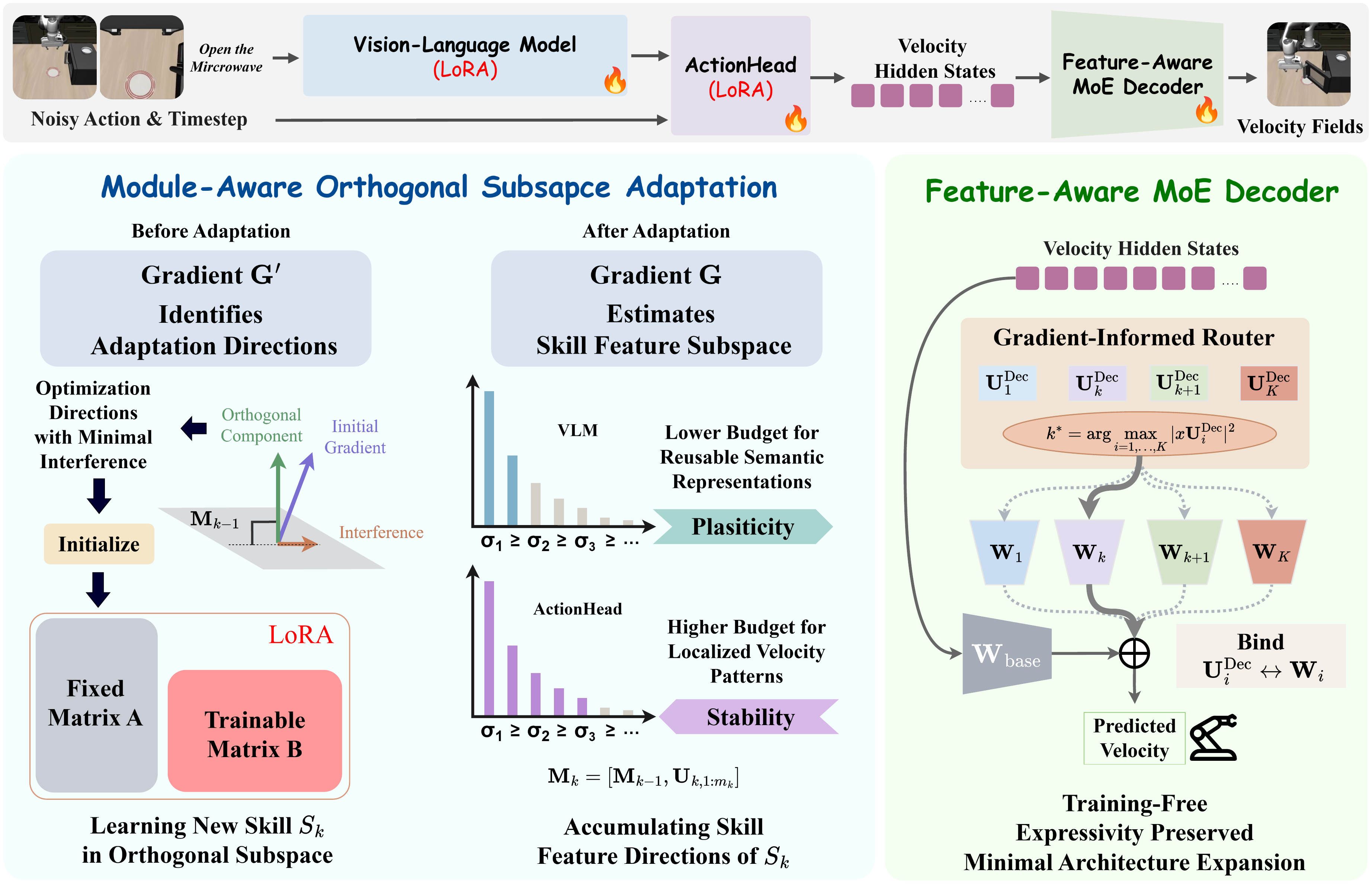}
    \caption{
        Overview of OrthoSkillVLA.
        For the VLM and ActionHead, OrthoSkillVLA constrains new skill learning to gradient-informed orthogonal subspaces with module-specific budgets, preserving reusable semantic capacity while protecting localized velocity patterns.
        For the final velocity decoder, it allocates a compact expert for each skill, and selects experts through a training-free projection router based on feature-space affinity.
    }
    \label{fig:1_framework}
\end{figure}

\section{Preliminary}

\paragraph{Continual Skill Learning for Pretrained VLA.} We consider a replay-free continual skill learning setting where an embodied agent sequentially masters a repertoire of skills $\mathcal{S}=\{S^1,\dots,S^K\}$ without access to historical datasets.
At stage $k$, the model only accesses demonstrations $D^k$ of the current skill $S^k$, and is expected to acquire $S^k$ while preserving its performance on prior skills $S^{<k}$.
During inference, the policy should execute all learned skills without relying on oracle skill identities.
As the underlying policy, we consider a pretrained VLA model composed of a Vision-Language Model (VLM) for semantic feature extraction, an ActionHead for predicting latent velocity states, and a final velocity decoder that maps these representations to physical velocity fields~\cite{intelligence2025pi05,zheng2025xvla}.

\paragraph{First-Order Interference and Gradient-Informed Subspaces.} To characterize interference between skills, consider a linear transformation $y=x\mathbf{W}+b$ inside the VLA model, where $\mathbf{W} \in \mathbb{R}^{d_{\text{in}} \times d_{\text{out}}}$.
After learning a new skill, the weight update $\Delta \mathbf{W}$ perturbs the output of previous skills by $x \Delta \mathbf{W}$.
A sufficient first-order condition for preserving the response on historical features is $x\Delta \mathbf{W} \approx 0, \forall x\in\mathcal{X}^{<k}$, indicating that updates for new skills should preferably lie in directions orthogonal to the feature subspace occupied by previous skills~\cite{liang2024inflora,luo2026keeplora}.

For a linear layer, the gradient of the loss with respect to $\mathbf{W}$ can be written as $\nabla_{\mathbf{W}} \mathcal{L}=x^\top \delta$, where $\delta$ is the back-propagated error.
Therefore, the accumulated gradients after learning a skill $S^k$ serve as a reliable proxy for the feature subspace occupied by $S^k$~\cite{saha2021gpm}.

\section{Method}
In this section, we present OrthoSkillVLA, a parameter-efficient continual learning framework for pretrained VLA models.
We first introduce the module-level subspace estimation and adaptation strategy to minimize the first-order interference in high-dimensional hidden layers of the VLA model.
Then we present the design of the Feature-Aware MoE Decoder and the training-free routing mechanism based on feature affinity to provide sufficient output flexibility in velocity decoding.
\subsection{Skill Subspace Estimation with Module-Aware Budgets}\label{sec:skill_subspace_estimation}

For each adapted layer, we maintain a compact orthonormal basis set $\mathbf{M}_{k-1} \in \mathbb{R}^{d_{\text{in}} \times \sum_{i=1}^{k-1} m_i}$ to span the feature subspace occupied by all previously learned skills, where $m_i$ is the number of basis vectors allocated for skill $S^i$.
Given the accumulated gradient $\mathbf{G}_{k}$ of the current skill, we project it onto the orthogonal complement of $\mathbf{M}_{k-1}$ to filter novel feature directions:
\begin{equation}\label{eq:project_gradient_after_skill}
    \hat{\mathbf{G}}_{k}=\mathbf{G}_{k} - \text{Proj}_{\mathbf{M}_{k-1}}(\mathbf{G}_{k})
    \quad \text{where} \quad
    \text{Proj}_{\mathbf{X}}(Y) = \mathbf{X} \mathbf{X}^\top Y,
\end{equation}
followed by Singular Value Decomposition (SVD) on the residual gradient: $\hat{\mathbf{G}}_k = \mathbf{U}_k \mathbf{\Sigma}_k \mathbf{V}_k^\top$.
Retaining all novel feature directions would aggressively consume the finite representational space, leaving restricted plasticity for future skill acquisition and unnecessary storage overhead.
To determine an appropriate $m_k$ for the current skill, an energy threshold $\epsilon \in (0, 1)$ is applied to guarantee that the filtered novel features $\mathbf{U}_{k, 1:m_k}$ reconstruct the current skill feature space with sufficient fidelity:
\begin{equation}\label{eq:energy_threshold}
    \frac{\| \text{Proj}_{\mathbf{U}_{k, 1:m_k}}(\hat{\mathbf{G}}_k) \|^2_F + \| \text{Proj}_{\mathbf{M}_{k-1}}(\mathbf{G}_k) \|^2_F}{\| \mathbf{G}_k \|^2_F} \geq \epsilon
\end{equation}
where $\|\cdot\|_F$ denotes the Frobenius norm.
Then the basis set is updated accumulatively: $\mathbf{M}_{k}=[\mathbf{M}_{k-1}, \mathbf{U}_{k,1:m_k}]$.

Acting as a subspace budget, a higher energy threshold leads to stronger retention of previously learned skills but reduced capacity for future skill acquisition.
The VLM and ActionHead play distinct roles in flow-matching VLA models.
Pretrained on diverse multimodal datasets, the VLM part maintains dense representations that require a relatively broader but reusable semantic subspace, making it vulnerable to capacity exhaustion.
The ActionHead, on the other hand, refines these broad semantics into specific velocity representations, rendering its dominant feature directions highly localized and sensitive to perturbations.
The representational heterogeneity between them necessitates differentiated capacity requirements during new skill acquisition.
We therefore adopt separate module-level budgets: a relatively relaxed threshold $\epsilon_{\text{VLM}}$ for the VLM to preserve long-term plasticity and a more stringent threshold $\epsilon_{\mathrm{Head}}$ for the ActionHead to ensure precise retention of velocity patterns.

\subsection{Gradient-Informed Orthogonal Low-Rank Adaptation}\label{sec:skill_subspace_adaptation}
At the beginning of acquiring skill $S^k$, the historical bases $\mathbf{M}_{k-1}$ summarize the feature space occupied by previously learned skills.
The optimization for $S^{k}$ is then constrained to the orthogonal complement of $\mathbf{M}_{k-1}$ in a parameter-efficient manner.
For each adapted linear layer, with the parameter update parameterized as $\Delta \mathbf{W}_{k}=\mathbf{A}\mathbf{B}$, its column space is bounded by the down-projection matrix $\mathbf{A}$.
Therefore, before fine-tuning on $S^{k}$, we compute the initial gradient $\mathbf{G}^{\prime}_{k}$ and project it away from the historical bases:
\begin{equation}\label{eq:project_gradient_before_skill}
    \hat{\mathbf{G}}_{k}^{\prime} 
    = \mathbf{G}_{k}^{\prime} - \text{Proj}_{\mathbf{M}_{k-1}}(\mathbf{G}_{k}^{\prime})
\end{equation}
To maximize the plasticity for the current skill, we initialize $\mathbf{A}$ with the top-$r$ left singular vectors of $\hat{\mathbf{G}}_{k}^{\prime}$.
During fine-tuning for skill $S^{k}$, only the up-projection matrix $\mathbf{B}$ is optimized while $\mathbf{A}$ remains frozen, ensuring that all updates are confined to the orthogonal subspace.
This mechanism allows the VLA model to sequentially integrate diverse velocity patterns while minimizing interference with previously learned skills.
After learning $S^k$, the low-rank adapters are merged into the base model and then perform subspace estimation as described in Section~\ref{sec:skill_subspace_estimation}.

\subsection{Feature-Aware MoE Velocity Decoder}\label{sec:feature_aware_moe_decoder}
In VLA models, the velocity decoder $\mathbf{W}_{\mathrm{base}} \in \mathbb{R}^{d_{\mathrm{in}}\times d_{\mathrm{action}}}$ serves as the final readout from latent representations to executable velocity fields.
Unlike hidden layers, whose representations can be further transformed by subsequent modules, this layer directly determines the generated trajectory precision.
Freezing the decoder would avoid disrupted velocity mappings, but create an optimization bottleneck.
In this case, the optimization burden of fitting heterogeneous velocity fields will be shifted onto the preceding LoRA adapters.
Crucially, these adapters are themselves confined to orthogonal subspaces to ensure stability.
Forcing them to compensate for a rigid output layer further exacerbates the tension between stability and plasticity.

To provide the necessary flexibility of learning velocity mappings, we introduce a Feature-Aware Mixture-of-Experts Decoder.
For each new skill $S^k$, we allocate a lightweight skill-specific expert $\mathbf{W}_k$, and freeze the pretrained decoder:
\begin{equation}\label{eq:moe_decoder}
    v_k = x_k (\mathbf{W}_{\mathrm{base}} + \mathbf{W}_k)
\end{equation}
where $x_k$ is the latent representation and $v_k$ is the predicted velocity.
By freezing $\mathbf{W}_{\mathrm{base}}$ and exclusively optimizing the skill-specific $\mathbf{W}_k$, the behavioral patterns of diverse skills are captured in isolated parameter spaces, providing the output-layer expressivity where it is most needed.

A challenge in such MoE architectures is the routing mechanism during inference.
Rather than introducing a complex trainable router that might itself be prone to forgetting, we propose a training-free gradient-informed routing strategy based on feature-space affinity.
According to Section~\ref{sec:skill_subspace_estimation}, after learning skill $S^k$, the accumulated gradient of $\mathbf{W}_k$ provides a reliable approximation of the space occupied by the velocity representation $x_k$.
We consolidate the left singular vectors of $\mathbf{G}_k^{\mathrm{Dec}}$ with the corresponding expert $\mathbf{W}_k$ inside the model to support training-free routing.
As a result, a test representation $x$ belonging to skill $S^k$ will naturally yield the highest projection magnitude onto its corresponding basis $\mathbf{U}_k^{\mathrm{Dec}}$.
Consequently, the expert selection is determined by:
\begin{equation}
    k^* = \arg\max_{i=1, \dots, K} \| x \mathbf{U}_i^{\mathrm{Dec}} \|^2
\end{equation}

\begin{algorithm}[t]
    \caption{OrthoSkillVLA: Continual Skill Learning Framework}
    \label{alg:orthoskillvla}
    \begin{algorithmic}[1]
        \State \textbf{Input:} Skills $\mathcal{S} = \{S^1, \dots, S^K\}$, pretrained VLA model weights $\theta_0$, adapted layers $\mathcal{L}$, module-level energy threshold $\epsilon_{\text{VLM}}$ and $\epsilon_{\text{Head}}$, LoRA rank $r$.
        \State \textbf{Initialize:}  Current VLA model from $\theta_0$; Historical bases $\{\mathbf{M}_0^\ell\}_{\ell\in\mathcal{L}}\leftarrow\emptyset$.
        \For{each skill $S^k \in \mathcal{S}$}
            \State $\triangleright$ \textit{Orthogonal Skill Subspace Adaptation}
            \State Compute initial gradients $\{\mathbf{G}_{k}^{\prime,\ell}\}_{\ell\in\mathcal{L}}$ on dataset $D^k$ and project it to the orthogonal complement of $\{\mathbf{M}_{k-1}^{\ell}\}_{\ell\in\mathcal{L}}$ (Eq.~\ref{eq:project_gradient_before_skill})
            \State Initialize and freeze down-projection $\{\mathbf{A}_{k}^{\ell}\}_{\ell\in\mathcal{L}}$ with the top $r$ singular vectors
            \State Allocate velocity-decoder expert $\mathbf{W}_k$ and freeze $\mathbf{W}_{\mathrm{base}}$ (Eq.~\ref{eq:moe_decoder})
            \State Optimize up-projection $\{\mathbf{B}_{k}^{\ell}\}_{\ell\in\mathcal{L}}$ and expert $\mathbf{W}_k$
            \State Merge LoRA updates to obtain $\theta_k$ and keep $\mathbf{W}_k$ isolated

            \State $\triangleright$ \textit{Module-Aware Skill Subspace Estimation}
            \State Accumulate gradient $\{\mathbf{G}_{k}^{\ell}\}_{\ell\in\mathcal{L}}$ and $\mathbf{G}_k^{\mathrm{Dec}}$ over dataset $D^k$
            \State Filter novel features of current skill (Eq.~\ref{eq:project_gradient_after_skill})
            \State  Update historical bases $\{\mathbf{M}_{k}^{\ell}\}_{\ell\in\mathcal{L}}$ using module-level energy thresholds $\epsilon_{\mathrm{VLM}}$ and $\epsilon_{\mathrm{Head}}$ (Eq.~\ref{eq:energy_threshold})
            \State Bind decoder-level routing basis $\mathbf{U}_{k}^{\mathrm{Dec}}$ with expert $\mathbf{W}_k$
        \EndFor
    \end{algorithmic}
\end{algorithm}

Since this projection-based routing can be efficiently vectorized, and the decoder parameters constitute a negligible fraction of the total model size (less than $0.01\%$ in our implementation), the MoE Decoder introduces minimal inference latency and memory footprint, ensuring real-time deployability.
The training procedure is summarized in Algorithm \ref{alg:orthoskillvla}.

\section{Experiments}

\begin{figure}[t]
    \centering
    \includegraphics[width=0.9\textwidth]{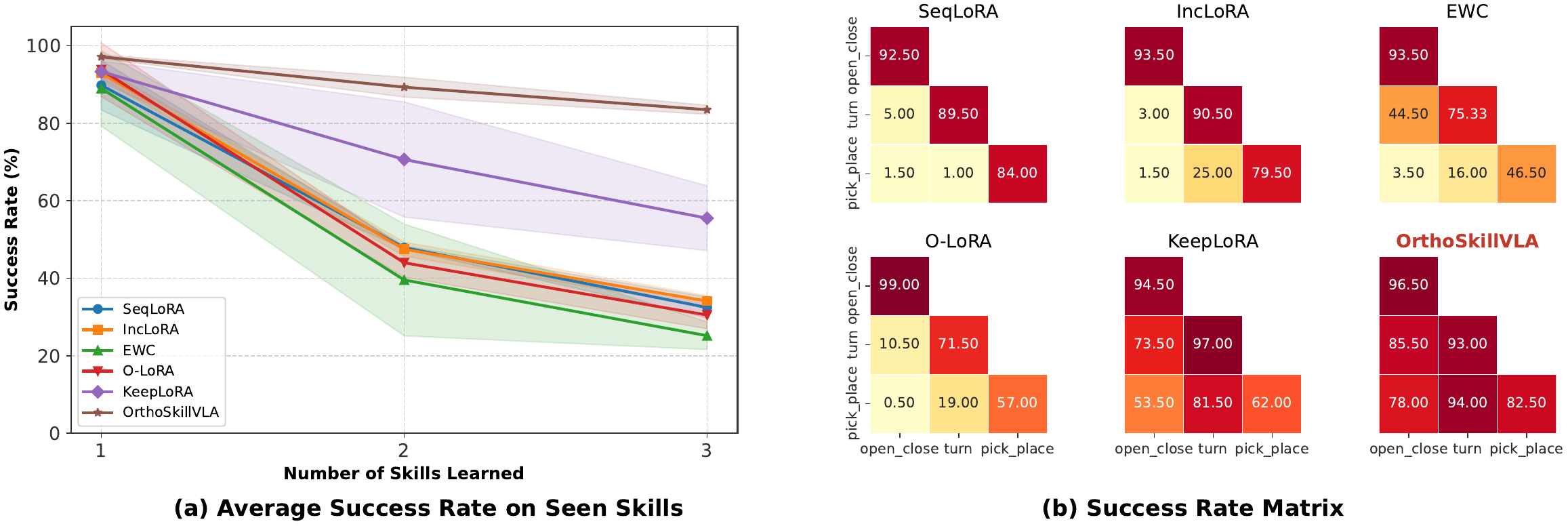}
    \caption{
        Simulation results under skill-incremental continual learning.
        (a) Average success rate on seen skills.
        (b) Skill-wise success matrices across learning phases.
    }
    \label{fig:sr_heatmap}
\end{figure}
In this section, we evaluate the continual skill learning capability of OrthoSkillVLA in simulated and real-world environments.
We adopt pretrained X-VLA~\cite{zheng2025xvla} as our base model due to its lightweight architecture ($\sim$ 0.9B) and effective adaptation capabilities proven on diverse simulation benchmarks and real-world applications.

\subsection{Simulated Experiments}
\paragraph{Experimental Setup.}
We perform our simulated evaluation on LIBERO~\cite{liu2023libero}, a lifelong robot learning benchmark.
While LIBERO offers diverse tasks, its original task-incremental protocols (Spatial, Object) primarily vary object identities or placements while maintaining similar motion patterns.
Such settings provide limited action distribution shifts, failing to fully expose the gradient conflicts in sequentially fitting heterogeneous velocity fields.

To this end, we introduce a more challenging skill-incremental evaluation protocol by reorganizing the LIBERO-100 suite into skill-specific task sets that emphasize pronounced discrepancies in action distributions.
Specifically, we define three skills, \textit{OpenClose}, \textit{Turn}, and \textit{PickPlace}, each comprising 2–-4 tasks to ensure substantial differences in motion patterns, comparable task counts, and the exclusion of compound skills.
To mitigate the influence of skill difficulties on our evaluation, we evaluate the average performance across three orderings.

To comprehensively evaluate OrthoSkillVLA, we compare it against representative replay-free continual learning and PEFT baselines:
\textbf{SeqLoRA}~\cite{wallis2021lora}, which sequentially fine-tunes LoRA adapters on each skill;
\textbf{IncLoRA}~\cite{wallis2021lora}, which reinitializes a new adapter for each skill without explicit interference mitigation;
\textbf{EWC}~\cite{kirkpatrick2017ewc}, which penalizes changes to parameters estimated to be important for prior skills;
\textbf{OLoRA}~\cite{wang2023olora}, which encourages LoRA updates for different skills to be orthogonal through an auxiliary loss;
and \textbf{KeepLoRA}~\cite{luo2026keeplora}, which fine-tunes LoRA adapters in an orthogonal subspace with a shared energy threshold for both the VLM and ActionHead.
The final velocity decoder is frozen to avoid corruption of previously learned velocity mappings.

Let $R_{i,j}$ denote the average success rate on skill $S^j$ after the model completes training on skill $S^i$. For a sequence of $K$ skills, to quantitatively assess continual learning capabilities, in addition to \textbf{Final SR}, where the final average success rate is evaluated, we also consider the following metrics:

\textbf{Forward Transfer (FWT)} measures the model's learning proficiency on newly introduced skills, defined as $\text{FWT} = \frac{1}{K} \sum_{k=1}^{K} R_{k,k}$.
\textbf{Negative Backward Transfer (NBT)} quantifies the severity of catastrophic forgetting to previously learned skills, defined as $\text{NBT} = \frac{1}{K-1} \sum_{k=1}^{K-1} \text{NBT}_k$, where $\text{NBT}_k = \frac{1}{K-k} \sum_{i=k+1}^{K} (R_{k,k} - R_{i,k})$.
\textbf{Area Under the Curve (AUC)} assesses the model's performance stability throughout the entire learning process, calculated as $\text{AUC} = \frac{1}{K} \sum_{k=1}^{K} \text{AUC}_k$, where $\text{AUC}_k = \frac{1}{K-k+1} \sum_{i=k}^{K} R_{i,k}$.

\begin{table}[t]
    \centering
    \caption{Comparison of continual learning metrics across different methods in the simulated skill-incremental setting. The best results are highlighted in \textbf{bold}.}
    \label{tab:sim_main_results_std}
    \renewcommand{\arraystretch}{1.2}
    \setlength{\tabcolsep}{4pt}
    \small
    \resizebox{\textwidth}{!}{
    \begin{tabular}{lcccc}
        \toprule
        \textbf{Method} & \textbf{FWT} $\uparrow$ & \textbf{NBT} $\downarrow$ & \textbf{AUC} $\uparrow$ & \textbf{Final SR}(\%) $\uparrow$ \\
        \midrule
        SeqLoRA~\cite{wallis2021lora} & $0.92 \pm  0.04$ & $0.89 \pm  0.05$ & $0.58 \pm  0.03$ & $32.44 \pm 3.31$\\
        IncLoRA~\cite{wallis2021lora} & $0.90 \pm  0.03$ & $0.83 \pm  0.06$ & $0.58 \pm  0.01$ & $34.11 \pm 1.70$\\
        EWC~\cite{kirkpatrick2017ewc} & $0.73 \pm  0.02$ & $0.68 \pm  0.04$ & $0.46 \pm  0.04$ & $25.17 \pm 4.38$\\
        OLoRA~\cite{wang2023olora}    & $0.88 \pm  0.10$ & $0.85 \pm  0.10$ & $0.54 \pm  0.07$ & $30.50 \pm 4.34$\\
        KeepLoRA~\cite{luo2026keeplora} & $0.90 \pm 0.05$ & $0.46 \pm 0.16$ & $0.72 \pm  0.02$ & $56.61 \pm 7.22$\\
        \textbf{OrthoSkillVLA} & \textbf{0.94 $\pm$ 0.03} & \textbf{0.13 $\pm$  0.06} & \textbf{0.88 $\pm$ 0.01} & \textbf{83.50 $\pm$ 1.42} \\
        \bottomrule
    \end{tabular}
    }
\end{table}

\paragraph{Experimental Results.}
Table~\ref{tab:sim_main_results_std} reports the main continual learning results averaged over three skill orderings.
SeqLoRA and IncLoRA achieve high FWT, but their large NBT and low Final SR indicate severe forgetting after sequential adaptation.
EWC mitigates forgetting partially, yet its strong regularization substantially weakens new-skill acquisition, leading to low FWT.
OLoRA provides only limited gains, suggesting that soft orthogonality is insufficient for fitting heterogeneous velocity fields.
KeepLoRA serves as the strongest subspace-constrained baseline, substantially improving retention over standard PEFT methods.
Nevertheless, its unified constraint remains sub-optimal for VLA architectures.
OrthoSkillVLA further reduces NBT from 0.46 to 0.13 and improves Final SR from 56.61\% to 83.50\%, demonstrating that Module-Aware Subspace Budgeting and Feature-Aware MoE decoder are both critical for continual skill learning.

\paragraph{Implementation Details.}
In our implementation, we use the AdamW optimizer with a batch size of 16 and a cosine annealing learning rate scheduler with a peak learning rate of $5 \times 10^{-5}$ after 1000 warmup steps.
For each individual skill, the acquisition process typically requires approximately 40,000 steps to reach convergence.
We adopt separated thresholds with $\epsilon_{\text{VLM}}=0.99$ and $\epsilon_{\text{Head}}=0.9999$ and defer the analysis to Section~\ref{sec:ablation_energy_strategy}.

\begin{table}[t]
    \centering
    \caption{
        Real-world success rates during continual skill learning phases for 20 trials. We compare the KeepLoRA~\cite{luo2026keeplora} baseline and OrthoSkillVLA.
    }
    \label{tab:real_main_results}
    \renewcommand{\arraystretch}{1.1}
    \setlength{\tabcolsep}{6pt}
    \begin{tabular}{l cccc cccc} 
        \toprule
        & \multicolumn{4}{c}{KeepLoRA~\cite{luo2026keeplora}} & \multicolumn{4}{c}{\textbf{OrthoSkillVLA}} \\
        \cmidrule(lr){2-5} \cmidrule(lr){6-9}
        Training Phase & Flip & Pick & Push & Press & Flip & Pick & Push & Press \\
        \midrule
        Phase 1: Learn \textit{Flip}       & 16 & --- & --- & --- & 19 & --- & --- & --- \\
        Phase 2: Learn \textit{Pick}       & 14 & 18  & --- & --- & 18 & 20 & --- & --- \\
        Phase 3: Learn \textit{Push}       & 11 & 13  & 15  & --- & 15 & 17 & 19 & --- \\
        Phase 4: Learn \textit{Press}      & 11 & 12  & 15  & 19  & \textbf{16} & \textbf{16} & \textbf{17} & \textbf{20} \\
        \bottomrule
    \end{tabular}
\end{table}
\subsection{Real-World Experiments}
\paragraph{Experimental Setup.}
To further validate the effectiveness of OrthoSkillVLA beyond simulated benchmarks, we conduct real-world experiments under a protocol analogous to the simulated setting.
We design four manipulation skills---\textit{Flip}, \textit{Pick}, \textit{Push}, and \textit{Press}---that span distinct action distributions.
\textit{Flip} emphasizes large end-effector orientation changes when reorienting a cup upright, while \textit{Pick} involves relatively stable pose control for placing an apple onto a plate.
\textit{Push} requires salient end-effector displacement to push a tripod onto a sponge, whereas \textit{Press} demands precise end-effector control to sequentially press two buttons without closing the fingers.

The real-world experiments are conducted on a 7-DoF xArm equipped with a 6-DoF Inspire dexterous hand and an Orbbec 336 wrist-mounted camera.
For each skill, we collect 50 expert demonstrations and the model is trained to predict the velocity fields in the normalized action space, with image observations resized to $224 \times 224$.
During evaluation, each skill is evaluated for 20 independent trials to compute the success rate.

\paragraph{Results and Analysis.}
Table~\ref{tab:real_main_results} reports the lower-triangular success matrix during real-world continual skill learning.
Compared with KeepLoRA, OrthoSkillVLA maintains more stable performance as new skills are sequentially introduced, while KeepLoRA exhibits more degradation on previously learned skills.
After completing all four skills, OrthoSkillVLA achieves an average final success rate of 86.25\%, outperforming KeepLoRA by 15.0\%.
The improvement mainly comes from better retention of earlier skills, while the later skills can still be acquired effectively.
These results demonstrate that OrthoSkillVLA better balances stability and plasticity on manipulation tasks with heterogeneous action distributions.

\section{Ablation Studies}
\subsection{Disentangling Module-Aware Budgeting and MoE Decoding}
To disentangle the effects of module-aware subspace budgeting and feature-aware MoE decoding, we compare four variants under the same protocol in our simulated experiments:
(i) \textit{Unified-Frozen}, which applies a single energy threshold to both the VLM and the ActionHead and freezes the velocity decoder, i.e., the KeepLoRA~\cite{luo2026keeplora} baseline;
(ii) \textit{Separated-Frozen}, which allocates relatively more budget to the ActionHead but still freezes the velocity decoder;
(iii) \textit{Unified-MoE}, which keeps the unified threshold but equips the model with the MoE velocity decoder;
and (iv) \textit{Separated-MoE}, the full OrthoSkillVLA.

As shown in Table~\ref{tab:disentangle_ablation}, when the velocity decoder is frozen, the separated threshold strategy provides slight improvements in NBT and Final SR, which is expected since a more stringent threshold for the ActionHead better preserves previously learned velocity representations.
In contrast, introducing the MoE decoder brings larger gains across all metrics, as skill-specific experts provide isolated velocity-mapping flexibility at the critical output stage.
Finally, the combination of both designs in OrthoSkillVLA achieves the best performance, reducing NBT from 0.46 to 0.13 and improving Final SR from 56.61\% to 83.50\%, which validates that module-aware budgeting and feature-aware MoE decoding are both critical for continual skill learning.
We further analyze the distinct roles of the VLM and ActionHead in VLA models and the routing mechanisms of the Feature-Aware MoE Decoder in Section~\ref{sec:ablation_energy_strategy} and Section~\ref{sec:ablation_routing}, respectively.

\begin{table}[t]
    \centering
    \caption{Ablation on module-aware budgeting and MoE velocity decoding.}
    \label{tab:disentangle_ablation}
    \setlength{\tabcolsep}{6pt}
    \resizebox{\textwidth}{!}{
    \begin{tabular}{llcccc}
        \toprule
        Threshold   & Decoder & FWT $\uparrow$ & NBT $\downarrow$ & AUC $\uparrow$ & Final SR (\%) $\uparrow$ \\
        \midrule
        Unified     & Frozen    & $0.90 \pm  0.05$ & $0.46 \pm 0.16$ & $0.72 \pm  0.02$ & $56.61 \pm  7.22$ \\
        Separated   & Frozen    & $0.91 \pm  0.06$ & $0.42 \pm 0.05$ & $0.75 \pm  0.04$ & $60.44 \pm  3.10$ \\
        Unified     & MoE       & $0.92 \pm  0.03$ & $0.33 \pm 0.15$ & $0.79 \pm  0.03$ & $67.94 \pm  4.40$ \\
        Separated   & MoE       & $0.94 \pm  0.03$ & $0.13 \pm 0.06$ & $0.88 \pm  0.01$ & $83.50 \pm  1.42$ \\
        \bottomrule
    \end{tabular}
    }
\end{table}
\begin{figure}[t]
    \centering
    \includegraphics[width=1.0\textwidth]{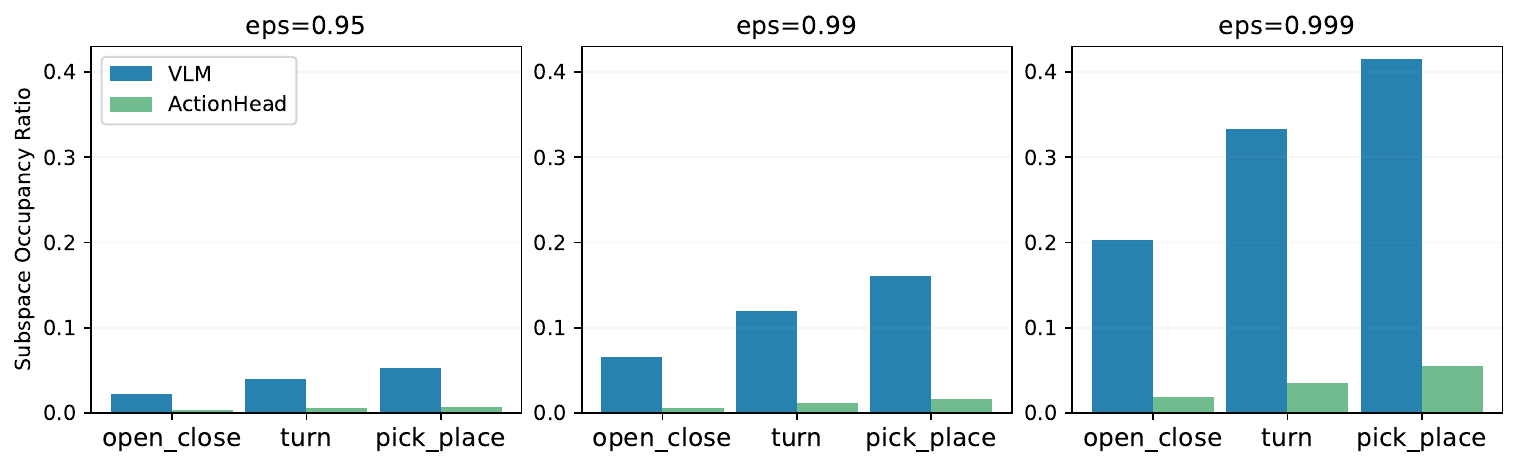}
    \caption{Evolution of SOR under varying unified energy thresholds.}
    \label{fig:subspace_occupancy_ratio}
\end{figure}

\subsection{Validating Module-Aware Budgeting Strategy}
\label{sec:ablation_energy_strategy}

To quantify the representation heterogeneity between the VLM and ActionHead, we measure the Subspace Occupancy Ratio (SOR), defined as the fraction of accumulated basis dimensions over the total input dimension of adapted layers after continual skill acquisition: $\text{SOR} = \frac{\sum_{\ell=1}^{L}\sum_{k=1}^{K} m_k^\ell}{\sum_{\ell=1}^{L} d_{\text{in}}^\ell}$.

As illustrated in Fig.~\ref{fig:subspace_occupancy_ratio}, applying a unified $\epsilon$ yields different SOR evolutions between VLM and ActionHead.
The VLM consistently occupies a much larger proportion of the subspace compared to the ActionHead.
This observed imbalance leads to a dilemma: knowledge retention in the ActionHead requires a stringent threshold that may aggressively exhaust the VLM's finite capacity, necessitating our module-aware budgeting strategy.

To further validate this strategy, we separately vary $\epsilon_{\mathrm{Head}}$ and $\epsilon_{\mathrm{VLM}}$ while fixing the other, and report both the final success rate and the final SOR to characterize the performance-capacity trade-off between them.
With $\epsilon_{\mathrm{VLM}}$ fixed to $0.99$ in Table~\ref{tab:ablation_threshold_fixed_vlm}, relaxing $\epsilon_{\mathrm{Head}}$ to $0.99$ causes a significant drop in Final SR, indicating that insufficient subspace budget for ActionHead fails to preserve the localized velocity patterns.
Further tightening $\epsilon_{\mathrm{Head}}$ to $0.9999$ substantially improves final performance to 84.83\% while keeping a moderate subspace occupancy of 23.07\%.
Under this sufficient budget for ActionHead, further allocating capacity to VLM brings a marginal gain of 0.67\% in Final SR but dramatically exhausts up to 53.73\% of the VLM's representational space, as shown in Table~\ref{tab:ablation_threshold_fixed_head}.

\begin{table}[t]
    \centering
    \begin{minipage}[t]{0.48\textwidth}
        \centering
        \caption{Ablation on the separated energy threshold strategy with \textbf{fixed} $\epsilon_{\text{VLM}}=0.99$.}
        \label{tab:ablation_threshold_fixed_vlm}
        \setlength{\tabcolsep}{2pt}
        \small
        \begin{tabular}{@{}lccc@{}}
            \toprule
            \multirow{2}{*}{$\epsilon_{\text{Head}}$} & \multirow{2}{*}{Final SR (\%)} & \multicolumn{2}{c}{Final SOR (\%)} \\
            \cmidrule(l){3-4}
            & & VLM & ActionHead \\
            \midrule
            0.99   & 59.83 & 22.57 &  2.17 \\
            0.999  & 80.67 & 21.09 &  7.57 \\
            0.9999 & 84.83 & 23.07 & 18.85 \\
            \bottomrule
        \end{tabular}    \end{minipage}%
    \hfill
    \begin{minipage}[t]{0.48\textwidth}
        \centering
        \caption{Ablation on the \textbf{separated} energy threshold strategy with \textbf{fixed} $\epsilon_{\text{Head}}=0.9999$.}
        \label{tab:ablation_threshold_fixed_head}
        \setlength{\tabcolsep}{2pt}
        \small
        \begin{tabular}{@{}lccc@{}}
            \toprule
            \multirow{2}{*}{$\epsilon_{\text{VLM}}$} & \multirow{2}{*}{Final SR (\%)} & \multicolumn{2}{c}{Final SOR (\%)} \\
            \cmidrule(l){3-4}
            & & VLM & ActionHead \\
            \midrule
            0.999 & 85.50 & 53.73 & 22.72 \\
            0.99  & 84.83 & 23.07 & 18.85 \\
            0.9   & 66.17 & 4.35  & 18.83 \\
            \bottomrule
        \end{tabular}
    \end{minipage}
\end{table}

\subsection{Analysis of Training-Free Routing Mechanism}
\label{sec:ablation_routing}

\begin{figure}[thbp]
    \centering
    \includegraphics[width=0.85\textwidth]{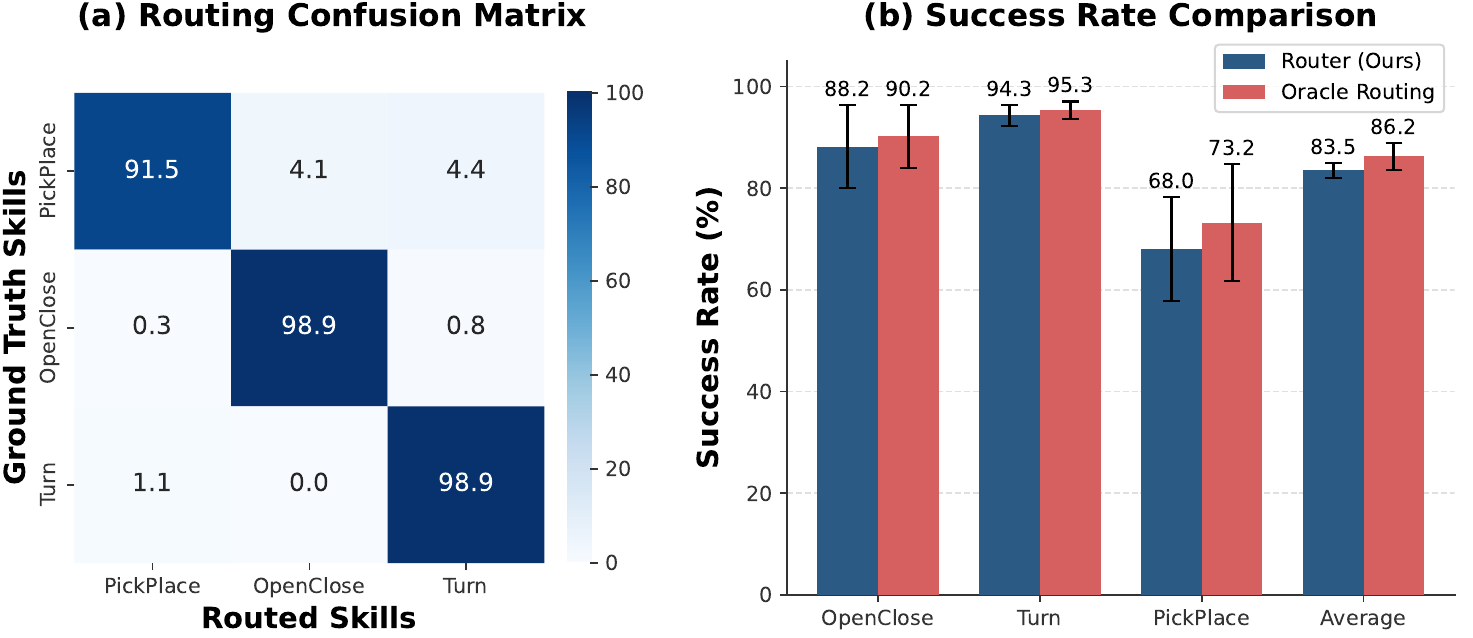}
    \caption{Evaluation of the training-free routing mechanism on routing accuracy \textbf{(a)} and success rate comparison with oracle routing \textbf{(b)} in our simulated benchmark.}
    \label{fig:routing_effectiveness}
\end{figure}

Regarding the routing mechanism, Fig.~\ref{fig:routing_effectiveness}(a) presents the routing confusion matrix, compiled across all numerical integration steps during inference.
The mechanism achieves highly accurate skill identification, yielding a reliable 98.9\% for \textit{OpenClose} and \textit{Turn}, and a robust 91.5\% for \textit{PickPlace}.
To quantify the impact of routing accuracy on final performance, Fig.~\ref{fig:routing_effectiveness}(b) compares our proposed mechanism against Oracle Routing, where the ground-truth skill identity is explicitly provided.

The results indicate that our training-free router performs closely to this theoretical upper bound, achieving an average success rate of 83.5\% compared to the oracle's 86.2\%.
The slight performance gap is primarily observed in the \textit{PickPlace} skill, which aligns with its relatively lower routing accuracy shown in the confusion matrix.
Overall, these findings validate the reliability of our feature-aware routing, demonstrating that it robustly distinguishes skills while preserving both inference efficiency and velocity expressivity.

\section{Conclusion}
Continual skill learning in pretrained VLA models requires preserving the representations and velocity mappings used by prior skills while adapting to diverse new skills.
We show that this challenge is shaped by two VLA-specific factors, namely the heterogeneous subspace-budget requirements of the VLM and ActionHead, and the output-stage expressivity bottleneck of the final velocity decoder.
Guided by these observations, OrthoSkillVLA combines module-aware subspace budgeting with a lightweight feature-aware decoder, and experiments validate that both module-specific protection and decoder flexibility are essential for stable continual skill acquisition.

\section{Acknowledgements}
This work was supported by the National Natural Science Foundation of China (Grant No 62476054) and the Southeast University Interdisciplinary Research Program for Young Scholars (2024FGC1007).

\nocite{*}

\bibliographystyle{splncs04}
\bibliography{references}

\clearpage
\appendix
\section{Details of Simulated Experiments}
\subsection{The Skill-Incremental Setting}
This work studies catastrophic forgetting in pretrained VLA models when they continually acquire diverse manipulation skills.
To better simulate the distributional shifts during continual skill learning, we reorganize LIBERO~\cite{liu2023libero} benchmark into a skill-incremental setting.
We start from LIBERO-100 and filter out compound tasks that involve multiple manipulation primitives, e.g. ``open the top drawer of the cabinet and put the bowl in it''.
The remaining 82 short-horizon tasks are then grouped into \emph{OpenClose}, \emph{PickPlace}, and \emph{Turn} according to their dominant behavioral patterns, with 8, 72, and 2 candidate tasks, respectively.
To avoid an evaluation dominated by the much larger \emph{PickPlace} group, we select a compact and more balanced subset while preserving diversity in scenes and interacted objects.
The final setting contains 4 \emph{OpenClose} tasks, 4 \emph{PickPlace} tasks, and 2 \emph{Turn} tasks.

\begin{table}[ht]
    \centering
    \caption{Task instructions of the three skills. See Figures~\ref{fig:open_close_tasks}--\ref{fig:turn_tasks} for execution snapshots.}
    \label{tab:selected_tasks}
    \setlength{\tabcolsep}{6pt} 
    \begin{tabular}{cl}
        \toprule
        \textbf{Skill} & \textbf{Task instruction} \\
        \midrule
        \multirow{4}{*}{\emph{OpenClose}} & close the bottom drawer of the cabinet \\
                                   & open the bottom drawer of the cabinet \\
                                   & close the top drawer of the cabinet \\
                                   & open the microwave \\
        \midrule
        \multirow{4}{*}{\emph{PickPlace}} & put the black bowl at the back on the plate \\
                                   & pick up the alphabet soup and put it in the basket \\
                                   & pick up the book and place it in the left compartment of the caddy \\
                                   & pick up the white mug and place it to the right of the caddy \\
        \midrule
        \multirow{2}{*}{\emph{Turn}} & turn on the stove \\
                                   & turn off the stove \\
        \bottomrule
    \end{tabular}
\end{table}

\begin{figure}[ht]
    \centering
    \includegraphics[width=\textwidth]{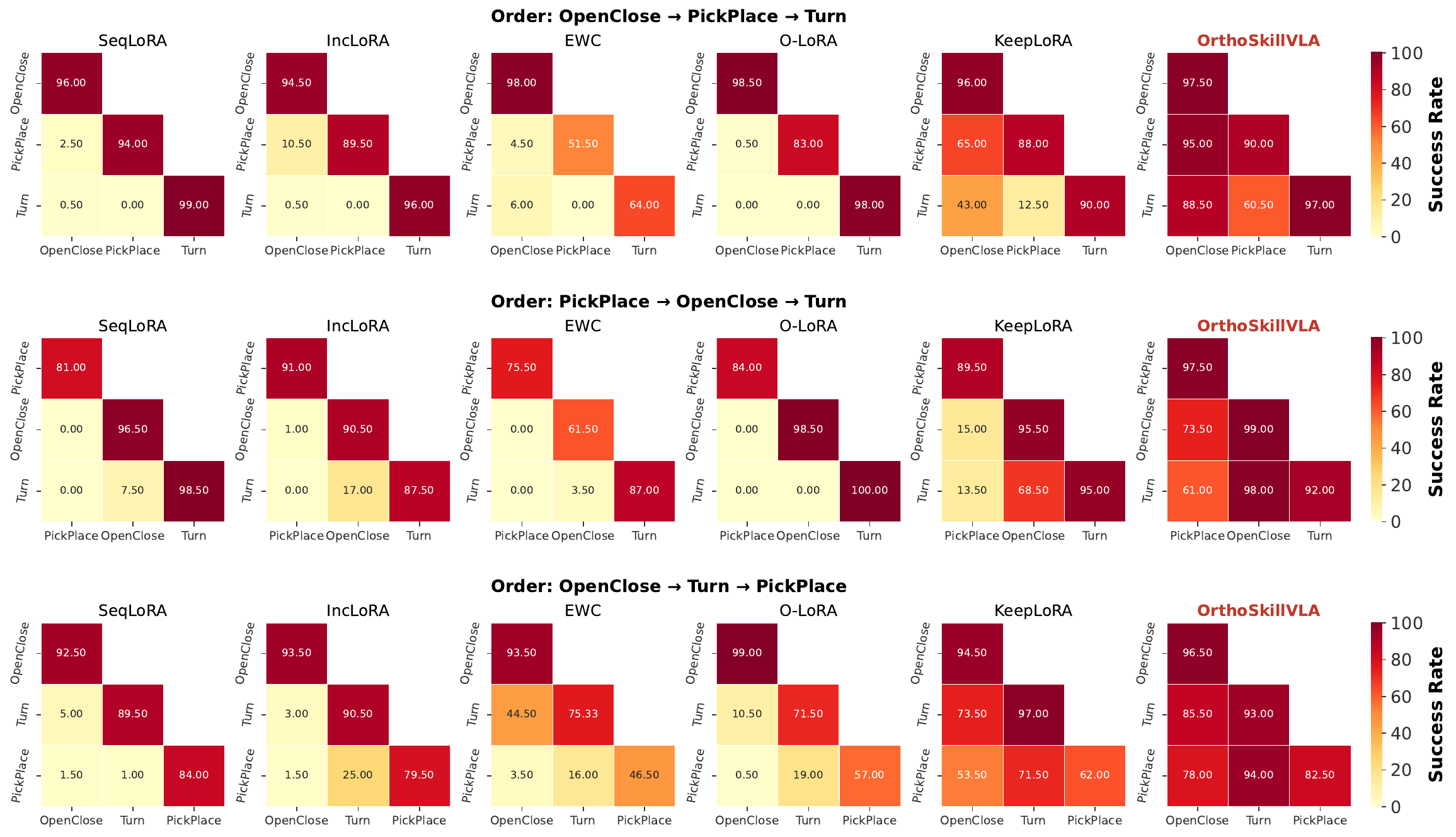}
    \caption{Qualitative comparison of model performance under the three skill orderings used in the simulated continual learning experiments.}
    \label{fig:sr_3_orders_rows}
\end{figure}

To mitigate order-specific bias, as shown in Table~\ref{tab:orders}, we evaluate all methods under three orderings and report the mean and standard deviation in the main paper.
During evaluation, each task is executed for 50 independent rollouts, and the success rate of a skill is computed by averaging over all tasks belonging to that skill.
A qualitative comparison of model performance under these three orderings is further provided in Fig.~\ref{fig:sr_3_orders_rows}.

\begin{table}[ht]
    \centering
    \caption{Skill orders used in the simulated continual learning experiments.}
    \label{tab:orders}
    \setlength{\tabcolsep}{10pt}
    \begin{tabular}{cc}
        \toprule
        Order ID & Skill orderings \\
        \midrule
        Ordering 1 & \emph{OpenClose} $\rightarrow$ \emph{PickPlace} $\rightarrow$ \emph{Turn} \\
        Ordering 2 & \emph{OpenClose} $\rightarrow$ \emph{Turn} $\rightarrow$ \emph{PickPlace} \\
        Ordering 3 & \emph{PickPlace} $\rightarrow$ \emph{OpenClose} $\rightarrow$ \emph{Turn} \\
        \bottomrule
    \end{tabular}
\end{table}

\subsection{More Implementation Details}
\paragraph{Starting checkpoint.} Following X-VLA~\cite{zheng2025xvla}, which absorbs embodiment heterogeneity with domain-specific soft prompts and action-related input/output projections, we treat LIBERO as a new embodiment domain before continual skill learning.
We extend the domain dimension for LIBERO by one and initialize the new entries with the average of existing pretrained domains.
Then we warm up only these newly introduced parameters on LIBERO-Goal, LIBERO-Spatial, and LIBERO-Object, with the shared backbone frozen.
The warm-up tasks are disjoint from our LIBERO-100 skill-incremental evaluation and serve only to obtain an aligned LIBERO-domain initialization.
All compared methods start from the same checkpoint.

\paragraph{Demonstrations.} We use the LIBERO datasets released with the X-VLA~\cite{zheng2025xvla} pretrained model and convert them into the LeRobot~\cite{cadene2026lerobot} format for training.
Each demonstration contains two visual observations from a wrist-view camera and a third-person camera.
Following the implementation of X-VLA, the effective action space for LIBERO control is represented as a 10-dimensional vector, consisting of 3 dimensions for end-effector translation, 6 dimensions for the 6D rotation representation~\cite{zhou2019continuity}, and 1 dimension for gripper control.
The remaining 10 action channels in the original X-VLA interface are reserved for bimanual embodiments and are not used in our experiments.

\paragraph{Architectural footprint analysis.} For the MoE velocity decoder, each newly acquired skill stores a skill-specific decoder expert together with its associated routing basis.
Let $d_h$ denote the hidden size, $d_a$ the action dimension, and $n_{\mathrm{basis}}$ the number of orthonormal vectors retained for routing.
The additional per-skill decoder state consists of the expert weights and the routing basis, resulting in $d_h d_a + d_h n_{\mathrm{basis}}$ stored floating-point values.
After learning $K$ skills, the total decoder-specific growth is therefore $O(K d_h(d_a+n_{\mathrm{basis}}))$, leaving the dominant VLM and ActionHead computation unchanged.
In our experiments, with the hidden size of $1024$ and the number of orthonormal vectors $n_{\mathrm{basis}}$ bounded by the action dimension $d_a=20$, the additional per-skill decoder state is approximately $0.0046\%$ of the 0.9B-parameter base model.

\paragraph{Low-rank adaptation target.} We apply orthogonal low-rank adaptation only to eligible high-dimensional linear layers in the VLM and ActionHead, where both the input and output dimensions are no smaller than the model hidden size of 1024.
 
We summarize the hyperparameters in Table~\ref{tab:training_config}.
The accompanying code repository provides access to both the common starting checkpoint and the LeRobot-formatted datasets.

\begin{table}[ht]
    \centering
    \caption{Training configuration and baseline-specific settings used in the simulated experiments.}
    \label{tab:training_config}
    \setlength{\tabcolsep}{6pt} 
    \begin{tabular}{ccc}
        \toprule
        Setting & Item & Value \\
        \midrule
        \multirow{9}{*}{Common}
        & Optimizer & AdamW \\
        & Peak learning rate & $5\times 10^{-5}$ \\
        & Warmup steps & 1000 \\
        & Scheduler & cosine annealing \\
        & Batch size & 16 \\
        & Precision & bfloat16 \\
        & Epochs per skill & 15 \\
        & Approx. steps per skill & $35{\sim}40$k \\
        & LoRA rank & 64 \\
        \midrule
        \multirow{2}{*}{EWC~\cite{kirkpatrick2017ewc}}
        & Fisher estimation samples & All training samples \\
        & Regularization weight & $\lambda = 30{,}000$ \\
        \midrule
        O-LoRA~\cite{wang2023olora}
        & Orthogonality Loss weight &  $\lambda=100$ \\
        \midrule
        KeepLoRA~\cite{luo2026keeplora}
        & Energy threshold & $\epsilon=0.999$ \\
        \bottomrule
    \end{tabular}
\end{table}

\subsection{The Effect of LoRA Rank}
To investigate the influence of the LoRA rank $r$, we evaluate our framework's continual learning performance across $r \in \{32, 64, 96\}$ in Table~\ref{tab:rank_metrics}.

As $r$ increases, FWT shows a mild upward trend, indicating that a larger adaptation subspace provides slightly stronger plasticity for newly introduced skills.
This is expected because the down-projection matrix $\mathbf{A}$ is initialized from the principal directions of the residual gradient, and a larger rank preserves more candidate update directions for fitting the current skill.
However, the benefit saturates quickly, suggesting that most useful adaptation directions are already captured by a moderate rank.
In contrast, an excessively large rank substantially degrades retention.
When $r$ increases to $96$, NBT rises from $0.13$ to $0.29$ and the final success rate drops from $83.50\%$ to $74.06\%$.
This indicates that including low-energy tail directions may introduce noisy or weakly constrained updates, which reduces the effectiveness of orthogonal adaptation and interferes with previously learned skills.
Overall, $r=64$ provides the best trade-off, achieving the highest AUC and final success rate while maintaining lower forgetting than both smaller and larger ranks.
We therefore use $r=64$ as the default setting in our main experiments.

\begin{table}[t]
    \centering
    \caption{Ablation on the impact of the LoRA rank.}
    \label{tab:rank_metrics}
    \begin{tabular*}{0.75\textwidth}{@{\extracolsep{\fill}} c c c c c }
        \toprule
        Rank & FWT $\uparrow$   & NBT $\downarrow$  & AUC $\uparrow$   & Final SR(\%) $\uparrow$ \\
        \midrule
        32   & 0.93 $\pm$ 0.01  & 0.17 $\pm$ 0.03   & 0.86 $\pm$ 0.02  & 79.50 $\pm$ 3.06 \\
        64   & 0.94 $\pm$ 0.03  & 0.13 $\pm$ 0.06   & 0.88 $\pm$ 0.01  & 83.50 $\pm$ 1.42 \\
        96   & 0.95 $\pm$ 0.01  & 0.29 $\pm$ 0.05   & 0.84 $\pm$ 0.02  & 74.06 $\pm$ 3.86 \\
        \bottomrule
    \end{tabular*}
\end{table}

\section{Details of Real-World Experiments}
For the real-world experiments, we evaluate our framework on four manipulation skills, including \emph{Flip}, \emph{Pick}, \emph{Push}, and \emph{Press}.
These skills cover distinct manipulation patterns, including large end-effector orientation changes, stable grasp-and-place behavior, salient end-effector displacement, and precise contact-rich control.

\begin{figure}[t]
    \centering
    \includegraphics[width=\textwidth]{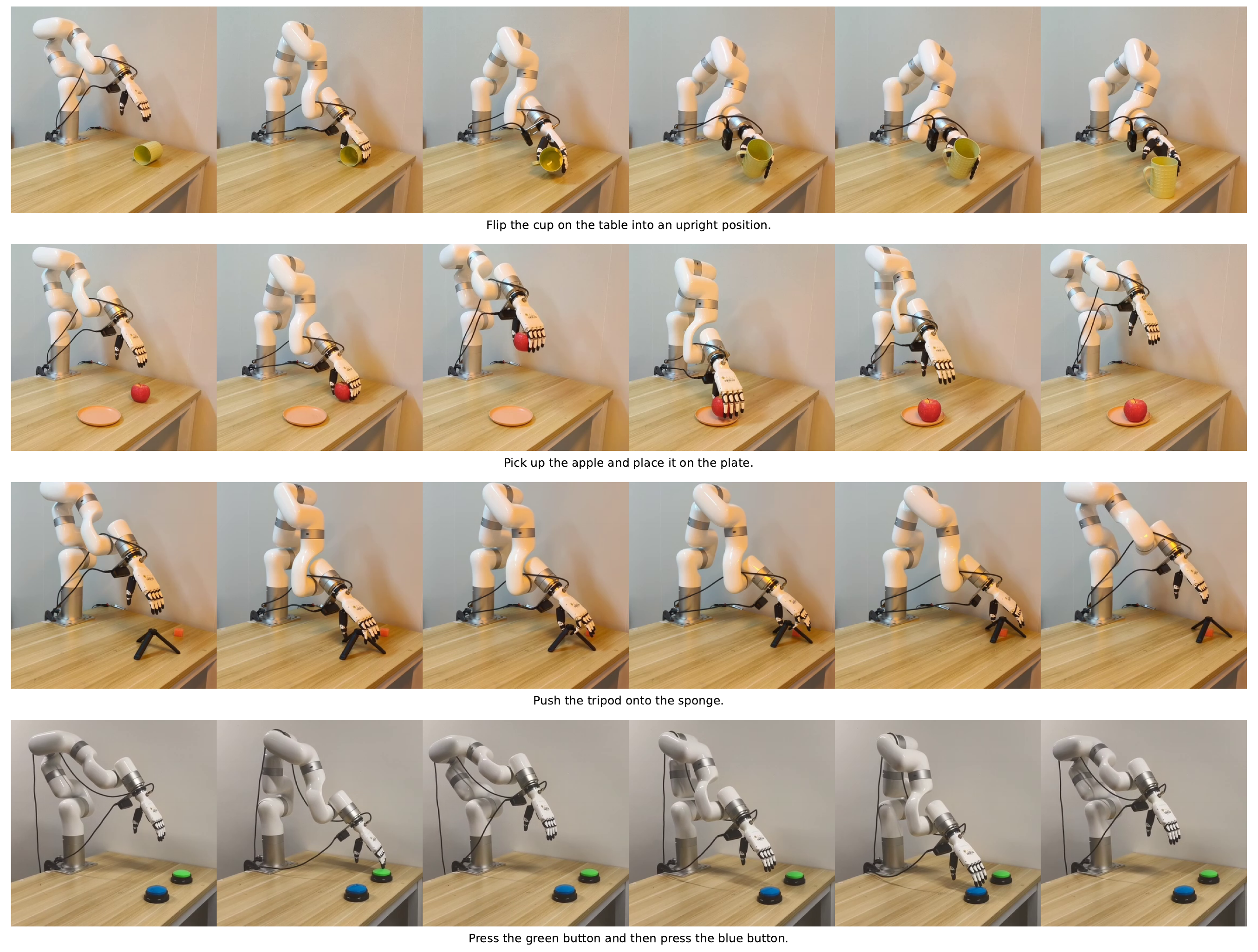}
    \caption{
        Visualization of the four real-world skills, with each row showing keyframes from an autonomous rollout of \emph{Flip}, \emph{Pick}, \emph{Push}, and \emph{Press}.
    }
    \label{fig:real_world_tasks}
\end{figure}

We collect 40 demonstrations for every skill.
Different from the end-effector in simulated experiments, the control command for the Inspire dexterous hand is 6-dimensional, resulting in a 15-dimensional action vector.
During training, the RGB observations recorded at $480 \times 480$ resolution are resized to $224 \times 224$ and the actions are normalized using min--max statistics computed from the collected demonstrations.
We provide execution snapshots to qualitatively visualize the four skills in Fig.~\ref{fig:real_world_tasks}.

\begin{figure}
    \centering
    \includegraphics[width=\textwidth]{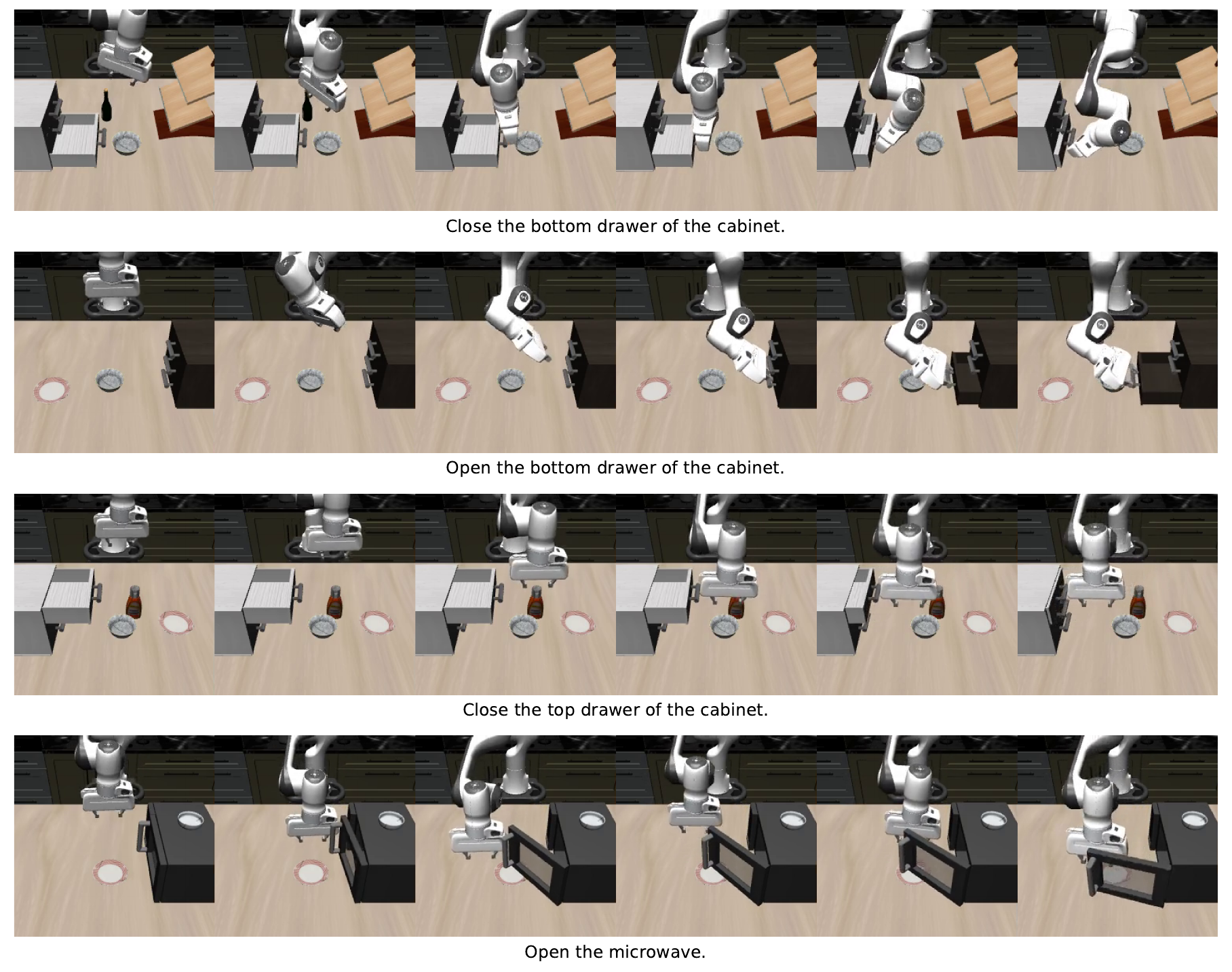}
    \caption{
        Visualization of the tasks in skill \emph{OpenClose}.
    }
    \label{fig:open_close_tasks}
\end{figure}

\begin{figure}
    \centering
    \includegraphics[width=\textwidth]{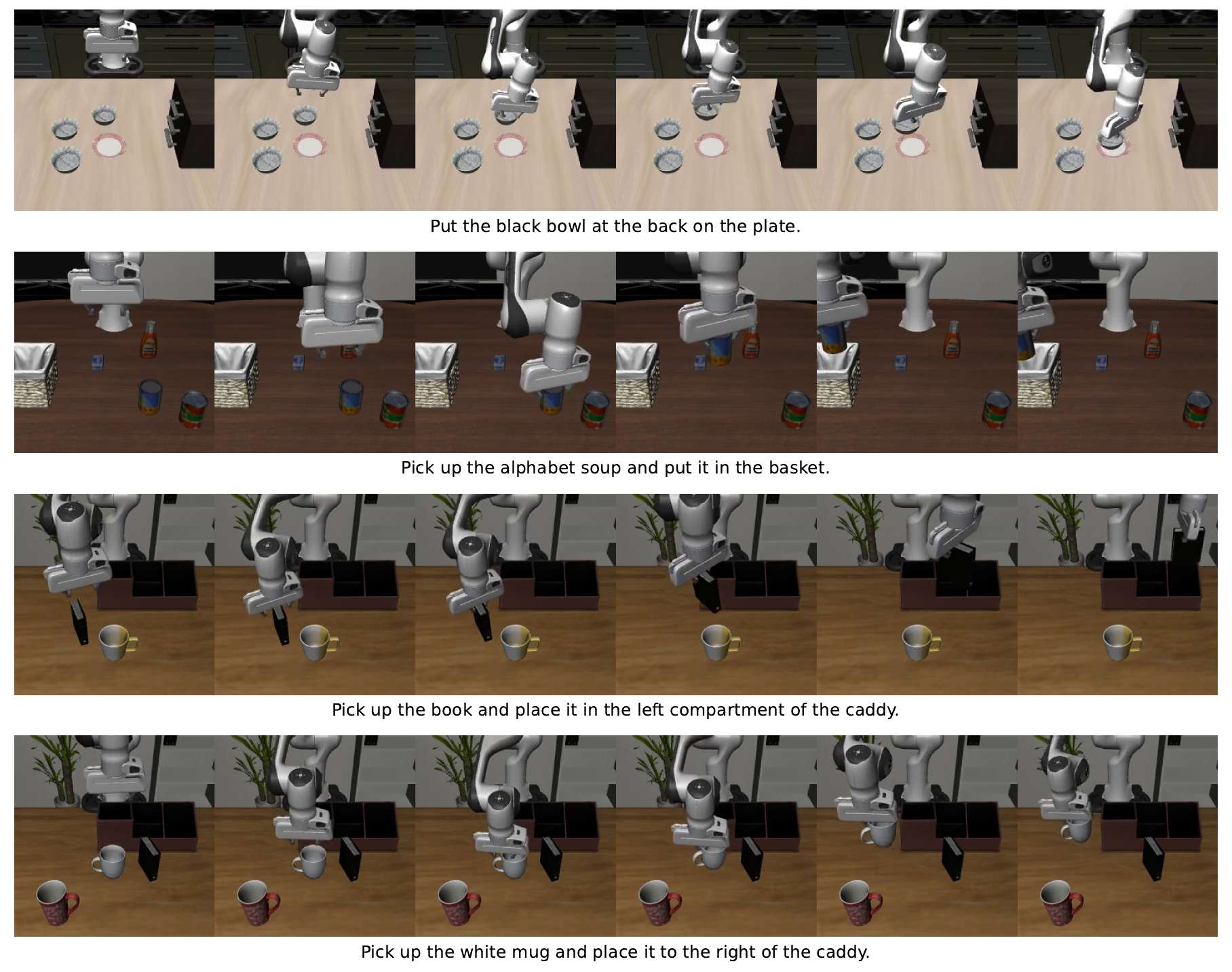}
    \caption{
        Visualization of the tasks in skill \emph{PickPlace}.
    }
    \label{fig:pick_place_tasks}
\end{figure}

\begin{figure}
    \centering
    \includegraphics[width=\textwidth]{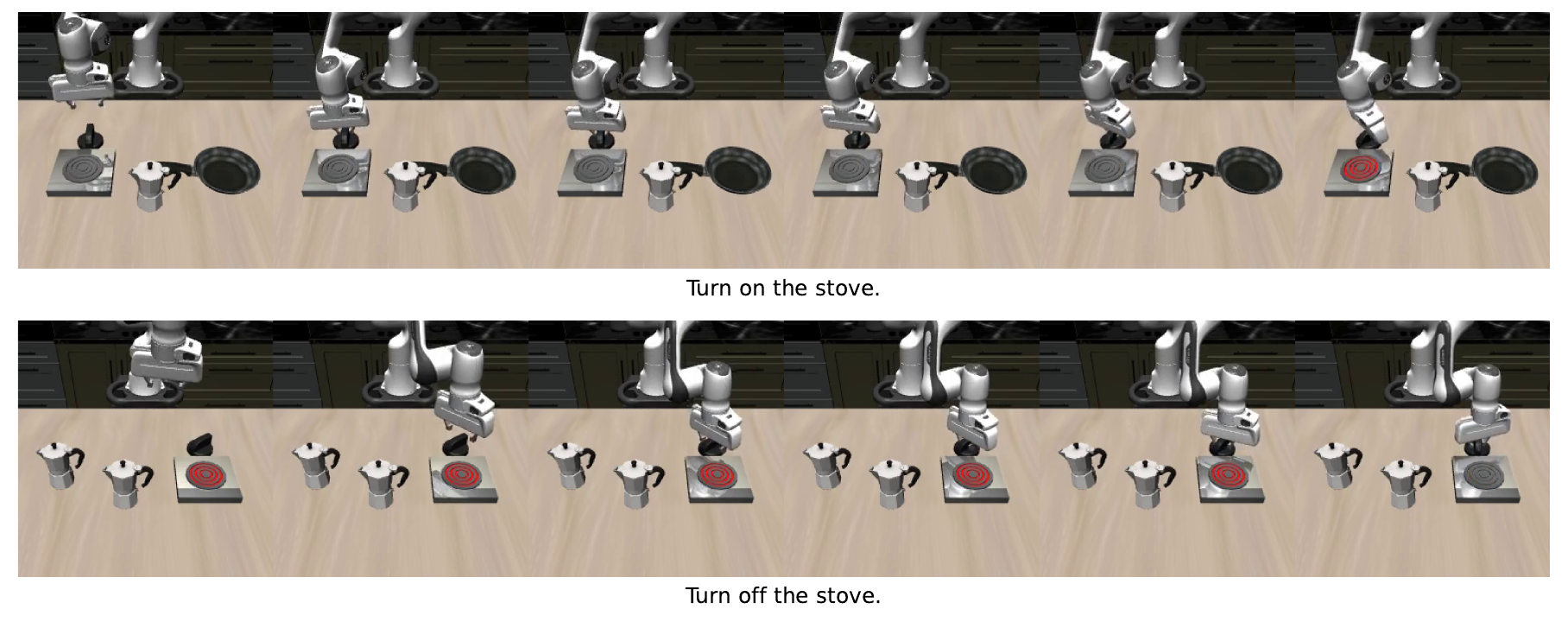}
    \caption{
        Visualization of the tasks in skill \emph{Turn}.
    }
    \label{fig:turn_tasks}
\end{figure}

\clearpage

\end{document}